# Simulation Techniques and Prosthetic Approach Towards Biologically Efficient Artificial Sense Organs- An Overview

Biswarup Neogi
ECE Dept., Sr. Lecturer
DIATM, Durgapur, India.
Email id:
biswarupneogi@gmail.com

Soumya Ghosal
IT Dept., Student
RCCIIT,Kolkata, India
Email Id:
soumyaghosal.2008@gmail.com

Soumyajit Mukherjee
CSE Dept., Student
SMIT,Kolkata, India
Email Id:
Soumyajitmukherjee.cs@gmail.com

Dr. Achintya Das
Professor & Head,ECE Dept.
Kalyani Government
Engineering College.
Email id:
achintya_das123@yahoo.co.in

Dr. D.N. Tibarewala
Director,
School of BioScience &
Engineering, Jadavpur University
Email id:
biomed_ju@yahoo.com

**Abstract.** An overview of the applications of control theory to prosthetic sense organs including the senses of vision, taste and odor is being presented in this paper. Simulation aspect nowadays has been the centre of research in the field of prosthesis. There have been various successful applications of prosthetic organs, in case of natural biological organs dis-functioning patients. Simulation aspects and control modeling are indispensible for knowing system performance, and to generate an original approach of artificial organs. This overview focuses mainly on control techniques, by far a theoretical overview and fusion of artificial sense organs trying to mimic the efficacies of biologically active sensory organs.

**1.Introduction.** One of the faculties by which the qualities of the external environment are appreciated that is sight, hearing, smell, taste or touch is called human sense. A collection of specialized cells (receptors) connected to the nervous system, that is capable of responding to a particular stimulus from outside or inside the body are the sense organs. As the lens helps a camera to track any object, similarly eye acts as an organ for human vision. A sensor tells us the identity of any object by sensing it; similarly nose acts as our body sensor, helping us to identify anything with its odor. Similarly tongue (the organ of taste) acts as a clinical sensor, and ear the organ of hearing works just as a microphone. All the human impulses coming from various sense organs, finally travels to the brain through the nervous system. Thus the brain acts as a control centre for different human actions. In this paper, a brief discussion is furnished on three artificial human sense organ control and modeling,i.e. eye, nose and tongue.

**2. Visual Acuity and Virtual reality leading towards prosthetic vision:** For centuries, restoration of proper eyesight to the blind has come to be considered nothing less than a miracle. Today, the resources are unlimited. Extensive researches and studies conducted today bear the promise of making restitution of sight a real medicinal achievement, tomorrow. Affordable for the common man, restoring eyesight would no longer come with the uncertainty of success. Pursuing that goal, leading researchers from the fields of ophthalmology, biology and computer engineering have united their efforts. This special report on bio-electronic vision, would reveal how for this field has been successfully explored and how much it is yet to be accomplished. Proper reception of prosthetic vision banks heavily upon the ability of the recipients in forming functional information from such vision. Several of the aforesaid critical factors were tested in a visual acuity study under virtual-reality replication of prosthetic vision using the Landolt C optotype. Fifteen subjects with normal sight were examined for many sessions. With regression models, the prime learning aspects were tested [1]. Learning was greatly directed toward a vital range of optotype sizes, and the subjects generally lacked ability in identification of the closed optotype (a Landolt C without any gap, creating a closed annulus). Preparation for implant receptors should aim these vital sizes as well as the closed optotype to eliminate the limitations of visual comprehension. Though no firm evidence can be provided that image processing influenced overall learning, personal preferences of the subjects varied widely. Several reports of successful clinical trials of visual neuro prosthesis prototypes indicate towards a major achievement made in the restoration of vision to the blind. The primary objective is to substitute the visual neural signaling with artificial signals reproduced by electrical stimulations. The success of the trials lies in eliciting spots of light, also termed as phosphenes, in the visual space, in correspondence to the application of electric current [2-9]. Drawing inspirations from these, many investigators have further researched simulated prosthetic vision with subjects having normal eyesight. Attempts are being made to explore more the extent of its capabilities and thereby gain more insight. The performance results received on the simulated prosthetic vision on visual acuity, object identification, ,hand-eye synchronization, fast reading ability [10–14], maze navigation [15] and object tracking [16][17], are optimistic enough that boost experimental activities in future. Learning is the core of this paper. Quite unfortunately, and also surprisingly, learning has gone largely unnoticed in the literature of prosthetic vision. Plasticity of the cortex aids learning to a large extent. The phenomenal success with the cochlear replacements for the hearing impaired is based upon plastic alterations in the auditory cortex [18]. Learning of visual assignments and the cortical systems underlying such procedures have also been well researched, revealing a visual cortex [19–21] with high adaptability. Furthermore, it is also established that the visual cortex of blind subject becomes accustomed to verbal memory and Braille reading tasks [22]. It is expected that the visual cortex would adapt to the "artificial" neural signals. The same has been illustrated in the auditory cortex of cats [23][24].These works attempts to look into the learning of prosthetic vision through virtual-reality simulation with human subjects having normal eyesight. Learning based on perseverance is mainly of two types: fast learning which depends on sensory units for executing the task, and slow learning where the cortical neurons and perceptual modules [25][26] are interconnected, thus strengthening the memory of fast learning into a more lasting memory. Since the early time the development in monosyllabic word identification by cochlear embedded patients can be assigned to speech recognition, as stated in[27]. Prosthetic vision can be said to be analogous to Uttal"s visual technique [28] and „vontour integration [29], other than the absence of Gabor function in phosphene elements as studied in visual psychophysics [30]. Its just a matter of time now after which guide dogs, and walking canes formerly used by blind people will be of no use to them as stated by Dobelle [31]. Proving him right, Humayun *et all* [32][33] and Veraart *et all* [34] showed that suitable electrical current wave form if passed through undamaged part of human eye can evoke rounded light spots called phosphenes. Many earlier works on prosthetic vision such as visual sharpness and speed of reading shown by Cha *et all* [35-37], Hayes *et all* [38]and Thompson *et all* [39] came up with ideas of added object recognition and facial recognition respectively gave us a vivid idea of prosthetic vision technique. The number and density of phosphene [40] gave us a new edge towards prosthetic vision analysis. Contracting information from every image frame in a limited phosphene number with the help of an algorithm was stated by Hallum *et all* [41], which described the task of tracing an object under simulated prosthetic vision. Spatial frequency component of Landolt ring and E-optotype discussed by Bondarko and Denilova [42] stated that lower frequency component is almost half of higher frequency component.

**3. Fuzzy neural network germinating Bio-electric prosthetic nose:** The artificial nose concept was around through the years and was developed by various researchers in this field till recent dates. This prosthetic device has been developed to grasp odors, vapors and gases automatically. Generally the constitution of artificial nose requires a sensor system and a pattern recognition system .An order-reactive polymer sensor, generating pattern of resistance make the classification of the odorant stimulus [43].Also the use of Artificial Neural Network produces proving results in Time Delay Neural Network [44]. However, hybrid application, as the use of Wavelet Analysis, the use of neuro-fuzzy networks and extraction of classification rules of the sensors of the artificial nose help to build this device[45-47].The optical gas sensors also demand large applications in this prosthetic nose field[48]. The optical technique based on UV-Vis spectroscopy has been described as an efficient tool to obtain the sensing response of these materials to various volatile organic compounds (VOCs) [49][50], by its measurement of the changes in spectra. Computing techniques have been adapted to solve the classification problems, KNN is one of them [51].From the past several years, development on the hybrid system for odor detection based on the olfactory organs of insects was going on [52][53]. Recordings from 21 individual glomeruli of honeybees have recently been used for the classification of several odors using the technique of principal component analysis (PCA) [54]. Original sensor fusion method on the basis of the opinions of human smell and taste and the measurement data from artificial nose and taste sensors is obtained from[55].Whether, in the market the commercial uses of electronic noses have already been started[56- [58],similar concepts for analysis of liquid is described, but for the tasting sense the terms „electronic tongue" or „taste sensor" have been used[59-62].To implement the pattern recognition system of artificial nose, studies have been done on the several types of ANN. The models like Multilayer Percept (MIPS) with, respectively, back propagation (BP), resilient back propagation (Rprop) and tabu search (TS), and networks with radial basis function (RBF networks) are analyzed [63-68].This overview paper on artificial nose cannot be flourished without the help of [69-72].

**4. Neural Network acting as an aid for the test receptors of an Electronic Artificial Tongue:** A muscular organ attached to the floor of human mouth, covered by a mucous membrane having minute projections (papillae) on its surface is the Tongue. It helps in manipulation during mastication and swallowing. Other than these, it is the primary organ of taste and also played an important role in producing speech [73]. Test has mainly five descriptors (salt, sour, sweet, bitter, umami) and for hot and cold [74]. Looking at the importance and usefulness of tongue many researchers aim for the development of sensory system related Electronic Tongue (E Tongue) which will mimic human sensory tongue both in structure and efficiency. An electronic tongue or „taste sensors" including arrays of ion selective and non-selective electrodes must be capable of detecting any chemical substance. Nowadays chemical/biological sensors for liquid analysis are widely accepted techniques in various research fields [75]. The first concept of the taste of sense was reported fifteen years ago [75][76]. For measuring the metal ions, present in river water, a calcogenide glass electrode based electronic tongue was modeled [76].

Followed by this, an electronic tongue aimed at analysis of beverages comprising of PVC and glass electrodes was introduced [77]. The membrane of electronic tongue comprised of eight kinds of lipid analogues. Based on lipid/polymer coated membrane, a taste sensory or electronic tongue was modeled [78] [79]. A kind of taste sensors are the resonant sensors. For detection of liquid species QCM is used, but the high oscillator damping gives rise to some electronic based requirements [80]. Not only this QCM also helps in miniaturizing the arrays [81] varying the sensitivity. Though the use of FPW (a type of resonant sensor) is known for long time but the first application of electronic tongue came much later [82] .But with utmost surprise and in contrary with all the electronic tongue models described till now, a completely different idea of electronic tongue came in 1997 [83]. Pulse Voltammetry used E. Tongue [84][85] consisting of six working electrodes of different metals, along with an auxiliary electrode and a reference electrode was then modeled. The voltammograms records were dependent on large amplitude pulse voltammetry (LAPV) [86]. Researchers of various ages were highly tilted towards solving the pattern recognition issue, in case of sense of taste with the help of an Artificial Neural Network (ANN) [87]. Their urge was more stimulated when Kyushu University scholars, Tokyo-Yamafuji came up with a taste transducer, using lipid membranes [88][89]. This group of Kyushu University came up with another prolific idea of multichannel taste sensor [90][91] consisting of eight different lipid membranes that converts the taste strength to electric potential and respond to different types of taste sensations. Electro physiological study on the thin lingual membrane of tongue has proved it to be a highly active transporting tissue and is shown in[92-95]. The reading difference between water readings and red wines in electronic tongue with the help of MLP networks is discussed in another work by Costa De Sousa et al. in 2002. The electro physiological studies of taste receptor cells that remain in cell collection of nearly 100 cells [96] together gives the sensation of salt [97]. Other than this the taste of beer and other food stuffs with the help of electronic sensing of electronic tongue was discussed in [98]. The reading difference between water readings and red wines in electronic tongue with the help of MLP networks was discussed in [99].

**5. Sound Source Localization Using Artificial Neural Network leading towards Prosthetic Ear:**

The artificial ear concept was around and developed through the years by various scientists and researchers in this field. The comparison and experiment between two promising classification techniques, non-windowed artificial neural networks (ANN) and hidden Markov models (HMM), with an artificial neural network using windowed input is shown in [100]. In spite of the huge use of hearing aids, it is responsible for different problems most often difficulty in hearing in environments with background noise [101]. These types of programmable hearing aids are currently available in market; however the setting can be changed manually [102]. Two of the most promising techniques introduced for audio classification were artificial neural networks (ANN) and hidden Markov models (HMM) [103]. A three-layer (one hidden layer) feed-forward perception with a variable number of hidden nodes type network was used in the work and it was trained using back-propagation [104]. Hidden Markov models were stochastic signal models, mainly based on Markov chains [105]. Although, many previous works have tested many possible features for this application [106], the most appropriate feature vector was dependent on both the classes [107] and the classifiers. Now, the aim of a system dealing with human-robot interaction is that the robot finds a person speaking in an area and goes to the speaker while the robot avoids static obstacles using mapping data obtained before. For this, the system acts like human especially when turning its head after someone calls. Now this intelligent and active system for Human-Robot interaction based on sound source localization has been discussed in this papers[108-110].The development of an 8-channel,real time, vibrotactile vocoder based on a TMS320C25 digital signal processing chip and by using FET spectral analysis technique is shown in [111].

## 6. Vibrotactile Sensing Elements for Artificial Skin Applications:

The skin is the largest stretched sense organ in human body. The replacement technique of artificial skin was from ancient times and the traditional method of replacement of skin was generated either by using skin from other parts of patients body or from a different person. Although, the disadvantages of the first technique was unavailability of ample skin and of the later technique was fear of being infected, lots of researches have been put on this artificial skin field and the consolidated studies about artificial skin have grown important eruditions about this field. The development of artificial skin surface ridges for incipient slip detection in pursuit of elucidating the mechanism of static friction sensing has been discussed by Yoji Yamada et all[112]. Now for the development of this plan it was determined to generate vibrotactile sensing capabilities on a skin tissue that has softness like human. Earlier, in a study about parallel change in the grip, the load forces was observed during precision grip of an object and the ratio between the two forces was adapted to result in the static friction coefficient between the finger skin surface and the object. This examination process was done by Johansson et al[113]. A tactile sensor system capable of detecting the incipiency of slip between an object and the sensor surface was proposed by Gaetano et al. by using the normal and shear stress information from arrays of PVDF transducers[114].The effectiveness of incipient slip detection more early using the peripheral slip signal from accelerometers which were mounted on a curved soft surface was discussed by Tremblay et al.[115].However, better shape and structure of the skin surface ridges with vibrotactile sensing elements could be designed and the dynamic response of human finger skin for tactile receptors focusing on the effect of epidermal ridges using FE analysis served as a background of the strategy of design[116].

**7.Conclusion:** During the last few decades, lot of clinical and research works have been carried out which points towards the future development of prosthetic studies. Our overview study is an effort to visualize all the recent works based on human prosthetic sense organs as much as possible. This overview is not intended to be an exhaustive survey on this topic, though a sincere effort has been made to cover all the recent works as much as possible and any omission of other works is purely unintentional. Future works aims at making smarter prosthesis, by better integrating the state of art- neuroscience with the state of art- engineering, medicine, computer and social science.

**REFERENCES:**

[1] Chen SC, Hallum LE, Lovell NH, Suaning GJ. Learning Prosthetic Vision: A Virtual-Reality Study. IEEE Transactions On Neural Systems And Rehabilitation Engineering. 2005;13:249-55.

[2] Humayun MS, Juan Jr ED, Dagnelie G, Greenberg RJ, Propst RH, Phillips DH. Visual perception elicited by electrical stimulation of retina in blind humans. Arch. Ophthalmol.. 1996; 114: 40–6.

[3] Humayun MS, Juan Jr ED, Weiland JD, Dagnelie G, Katona S, Greenberg R, Suzuki S. Patterned electrical stimulation of the human retina. Vis. Res. 1999; 2569–76.

[4] Humayun MS, Weiland JD, Fujii GY, Greenberg R, Williamson R, Little J., Mech B, Cimmarusti V, Van Boemel G, Dagnelie G. Visual perception in a blind subject with a chronic microelectronic retinal prosthesis.Vision Res. 2003; 43: 2573–81.

[5] Dobelle WH. Artificial Vision for the Blind by Connecting a Television Camera to the Visual Cortex. Amer. Soc. Artificial Internal Organs, 2000; 46: 3–9.

[6] Veraart C, Wanet-Defalgue M, Gerard B, Vanlierde A, Delbeke J. Pattern recognition with the optic nerve visual prosthesis. Artif. Organs, 2003; 27: 996–1002.

[7] Delbeke J, Oozeer M, Veraart C. Position size and luminosity of phosphenes generated by direct optic nerve stimulation. Vision Res.,2003; 43; 1091–102.

[8] Veraart C, Raftopoulos C, Mortimer JT, Delbeke J, Pins D. Visual sensations produced by optic nerve stimulation using an implanted self-sizing spiral cuff electrode. Brain Res., 1998; 813:181–6.

[9] Rizzo JF, Jensen RJ, Loewenstein J, Wyatt J. Unexpectedly small percepts evoked by epi-retinal electrical stimulation in blind humans. Invest. Ophthalmol .Vis. Sci., 2003; 44: 4207.

[10] Chen SC, Lovell NH, Suaning GJ. Effect of prosthetic vision acuity by filtering schemes, filter cut-off frequency and phosphene matrix: A virtual reality simulation. In 26th Annual Int. Conf. IEEE-EMBS, San Francisco, CA. 2004.

[11] Chen SC, Hallum LE, Lovell NH, Suaning GJ. Visual acuity measurement of prosthetic vision: A virtual-reality simulation study.J. Neural Eng., 2005; 2: S135–S145.

[12] Hayes JS, Yin VT, Piyathaisere D, Weiland JD, Humayun MS, Dagnelie G. Visually guided performance of simple tasks using simulated prosthetic vision. Artif. Organs, 2003; 27: 1016–28.

[13] Cha K, Horch K, Normann RA. Simulation of a phosphene based visual field: Visual acuity in a pixelized vision system. Ann. Biomed. Eng., 1992; 20: 439–49.

[14] Cha K, Horch K, Normann RA , Boman DK.Reading speed with a pixelized vision system. J. Opt. Soc. Amer. 1992; 9: 673–77.

[15] Cha K, Horch K, Normann RA. Mobility performance with a pixelized vision system.Vision Res., 1992; 32: 1367–72,

[16] Hallum LE, Taubman D, Suaning GJ, Morley J, Lovell NH. A filtering approach to artificial vision: A phosphene visual tracking task.

In IFMBE Proc. World Congr. Med. Phys. Biomed. Eng., Sydney, Australia, Aug. 24–29, 2003.

[17] Hallum LE, Suaning GJ, Taubman DS, Lovell NH. Simulated prosthetic visual fixation, saccade, and smooth pursuit. Vision Res., 2005; 45: 775–88.

[18] Lise GA, Eric T, Richard F. Imaging plasticity in cochlear implant patients. Audiol. Neuro-Otology, 2001: 6; 381–93.

[19] Karni A, Sagi D.The time course of learning a visual skill. Nature. 1993; 365: 250–52.

[20] Sagi D, Tanne D .Perceptual learning: Learning to see. Current Opinion Neurobiol., 1994; 4: 195–99.

[21] Tsodyks M, Adini Y, Sagi D.Associative learning in early vision. Neural Networks. 2004; 17: 823–32.

[22] Amedi A, Raz N, Pianka P, Malach R, Zohary E. Early visual cortex activation correlates with superior verbal memory performance in the blind. Nature Neurosci., 2003; 6: 758–66.

[23] Dinse HR, Godde B, Reuter G, Cords SM, Hilger T. Auditory cortical plasticity under operation: Reorganization of auditory cortex induced by electric cochlear stimulation reveals adaptation to altered sensory input statistics. *Speech Commun.*2003; 41: 201–19.

[24] Moore CM, Vollmer M, Leake PA, Snyder RL, Rebscher SJ.The effects of chronic intracochlear electrical stimulation on inferior colliculus spatial representation in adult deafened cats.*Hearing Res.*, 2002; 164: 82–96.

[25] Karni A.The acquisition of perceptual and motor skills: A memory system in the adult human cortex. *Cognitive Brain Res.*, 1996; 5: 39–48.

[26] Maquet P, Laureys S, Perrin F, Ruby P, Melchior G, Boly M, Dang Vu T, Desseilles M, Peigneux P. Festina lente: Evidences for

fast and slow learning processes and a role for sleep in human motor skill learning.In *Learning Memory*. 2003; 10: 237–39.

[27] Rubinstein JT, Miller CA. How do cochlear prostheses work ?. *Current Opinion Neurobiol.*, 1999; 9: 399–404.

[28] Uttal WR. Visual Form Detection in 3-Dimensional Space. Hillsdale, NJ: Lawrence Erlbaum, 1983.

[29] Field DJ, Hayes A, Hess RF. Contour integration by the human visual system: Evidence for a local association field. *Vision Res.*,

1993; 33: 173–93.

[30] Hess RF, Hayes A, Field DJ. Contour integration and cortical processing. *J. Physiology—Paris*. 2003; 97: 105–19.

[31] Dobelle WH. Artificial Vision for the Blind by Connecting a Television Camera to the Visual Cortex. *ASAIO Journal.*2000 ;46: 3-9.

[32] Humayun M, De Juan EJ, Dagnelie G, Greenberg R, Propst R, Phillips H. Visual perception elicited by electrical stimulation of retina in blind humans. *Arch Ophthalmol*, 1996; 114: 40–6.

[33] Humayun MS, Weiland JD, Fujii GY, Greenberg R, Williamson R, Little J, Mech B, Cimmarusti V, Van Boemel G, Dagnelie G, Visual perception in a blind subject with a chronic microelectronic retinal prosthesis. *Vision Research*, 2003 ; 43: 2573-81.

[34] Veraart C, Raftopoulos C, Mortimer JT, Delbeke J, Pins D, Michaux G, Vanlierde A, Parrini S, Wanet-Defalque MC. Visual sensations produced by optic nerve stimulation using an implanted self-sizing spiral cuff electrode. *Brain Research*, 1998 ; 813: 181-86.

[35] Cha K, Horch K, Normann R. Mobility performance with a pixelised vision system. *Vision Res*. 1992; 32: 1367-72.

[36] Cha K, Horch K, Normann R. Simulation of a phosphene-based visual field: Visual acuity in a pixelized vision system. *Ann Biomed Eng*. 1992; 20: 439-49.

[37] Cha K, Horch K, Normann R. Reading speed with a pixelized vision system. *J Op Soc Am*. 1992; 9: 673-77.

[38] Hayes JS, Yin VT, Piyathaisere D, Weiland JD, Humayun MS, Dagnelie G. Visually Guided Performance of Simple Tasks Using Simulated Prosthetic Vision. *Artificial Organs.*2003 ; 27: 1016-28.

[39] Thompson RW, Barnett GD, Humayun MS, Dagnelie G. Facial Recognition Using Simulated Prosthetic Pixelized Vision. *Invest. Ophthalmol. Vis. Sci.*2003; 44: 5035-42.

[40] Chen SC, Lovell NH , Suaning GJ. Effect on Prosthetic Vision Visual Acuity by Filtering Schemes, Filter Cut-off Frequency and Phosphene Matrix: A Virtual Reality Simulation: Proceedings of the 26th Annual International Conference of the IEEE EMBS San Francisco, CA, USA,2004; 4201-04.

[41] Hallum L, Suaning G, Taubman D, Lovell N. Simulated prosthetic visual fixation, saccade, and smooth pursuit; and the use of nontrivial image processing to effect improved prosthetic vision. *Vision Research*. 2004; accepted.

[42] Bondarko VM , Danilova MV. What spatial frequency do we use to detect the orientation of a Landolt C?. *Vision Research*. 1997; 37: 2153-56.

[43] Gardner JW, Hines EL. Porrern Analysis Techniques. Handbook of Biosensors and Electronic Noses: E. Kress-Rogers, editor. Medicine: Food *a d* the Environment. New York CRC Press; 1997.pp- 633452.

[44] Yamaraki A, Ludermir TB. Classification of Vintages of Wine by an Artificial Nosewith Neural Networks. Proceedings of the 8[th] international Conference on Neural Information Processing (ICONIP,2001),Shanghai. China, 2001; I: 184-87.

[45] Zanchettin C, Ludermir TB, Yamaraki A. Classification of Gases from the Petroliferous Industry by an Artificial Nose with Neural Network. Proceedings of the Joint 13th International Conference on Artificial Neural Networks and 10th International conference on Neural Informotion Processing , Istanbul. Turkey, 2003.

[46] Zanchettin C, Ludemir TB. Wavelet Filter for Noise Reduction and Signal Compression in an Artificial Nose. Proceedings of the Hybrid Intelligent system (HIS 2003), Melbourne, Australia. 2003; 907-16.

[47] Zanchettin C, Ludermir TB. A Neuro-Fuzzv Model Applied to Odor Recognition in an Artificial Nose. Proceedings of the Hybrid Intelligent system (HIS 2003). Melbourne. Australia. 2003; 917-26.

[48] Kladsomboon S, Puntheeranurak T, Pratontep S, Kerdcharoen T . An Artificial Nose Based on M-Porphyrin (M = Mg,Zn) Thin Film and Optical Spectroscopy. In 3rd IEEE International Nanoelectronics Conference,INEC 2010,Hong-Kong. 2010; 968-69.

[49] Uttiya S, Pratontep S, Bhanthumnavin W, Buntem R, Kerdcharoen T. Volatile organic compound sensor arrays based on zincphthalocyanine and zinc porphyrin thin films.In 2nd IEEE International Nanoelectronics Conference, INEC 2008, Shanghai. 2008; 618-23.

[50] Kladsomboon S, Pratontep S, Uttiya S, Kerdcharoen T. Alcohol gas sensors based on magnesium tetraphenyl porphyrins. In 2[nd] IEEE International Nanoelectronics Conference, INEC 2008,Shanghai. 2008; 585-88.

[51] Arshak K, Lyons GM, Cunniffe C, Harris J, Clifford S . A review of digital data acquisition hardware and software for a portable electronic nose. Sensor Review. 2003; 23: 332-44.

[52] Park KC, Ochieng SA, Zhu J, Baker TC. Odor discrimination using insect electroantennogram responses from an insect antennal array. Chemical Senses. 2002; 27: 343-52.

[53] Hetling JR, Myrick AJ, Park KC, Baker TC . Odor Discrimination Using a Hybrid-Device Olfactory Biosensor. Proceedings of the 1st International IEEE EMBS Conference on Neural Engineering, 2003.

[54] Galan RF, Sachse S, Galizia CG, Herz AVM. Odor-Driven Attractor Dynamics in the Antennal Lobe Allow for Simple and Rapid Olfactory Pattern Classification. Neural Computation, 2004; 16: 999-1012.

[55] Wide P, Winquist F, Bergsten P, Petriu EM. The Human-based multi-sensor fusion method for artificial nose and tongue sensor data. IEEE Instrumentation and Measurement Technology Conference St. Paul, Minnesota, USA, 1998;531-36.

[56] Alfa MOS, Tolouse, France. http://www.alpha-mos.com/

[57] Aroma Scan Inc., Crewe, U.K.

[58] Nordic Sensor Technologies, Linkoping, Sweden.

[59] Di Natale C, Davide F, Amico AD, Legin A, Rudinitskaya A, Selezenev BL, Vlasov Y. Applications of an electronic tongueto the environmental control. Technical digest of Eurosensors X, Leuven, Belgium. 1996;1345-48.

[60] Toko K. Taste sensor with global selectivity. Materials Science and Engineering C4 , 1996: 69-82.

[61] Legin A, Rudinitskaya A, Vlasov Y, Di Natale C, Davide F, "Amico AD. Tasting of beverages using an electronic tongue based on potentiometric sensor array. Technical digest of Eurosensors X, Leuven, Belgium. 1996; 427-30.

[62] Sasaki Y, Kanai Y, Ushida H, Katsube T. Highly sensitive taste sensor with a new differential LAPS method. Sensors and Actuators B 24-25. 1995; 819-22.

[63] Barbosa MS . Pattern Recognition of gases of petroleum based on RBF model" In: Proc. of VIII Brazilian Symposium on Neural Network. Los Alamitos: IEEE Computer Society. 2002;1:111.

[64] Rumelhart DE, Hinton GE, William RI. Learning internal representations by error propagation. Parallel Distributed Processing. D.E. Rumelhart and J.L.McClelland, editors. Cambridge. MIT Press: 1986;1**:** pp. 318-62.

[65] Santos MS . Artificial Nose And Data Analysis Using Multi Layer Perception. In: Data Mining, WIT Press, Computational Mechanics Publication. 1998.

[66] Yamazaki A, Ludermir TB. Neural Network Training wth Global Optimization Technique. International Journal of Intelligent Systems. 2003;13:77 - 86.

[67] Yamazaki A, De Souto MCP , Ludermir TB. Optimization of Neural Network Weight and architechtures for odor Recognition using Simulated annealing. In Proc. of the IEEE International Joint Conference on Neural Networks ,Honolulu. Hawaii. 2002;547-52.

[68] Yamazaki A, Ludermir TB , De Souto MCP. Global Optimization Methods for Designing and Training Neural Networks, In Proc. Of the VII Brazilian Symposium on Neural Networks, Brazil. 2002;130,135.

[69] Zanchettin C, Ludermir TB. Evolving Fuzzy Neural Networks Applied to Odor Recognition in an Artificial Nose. Neural Netw , Proceedings: 2004 IEEE International Joint Conference . 2004; 675-80.

[70] Myrick AJ, Baker TC, Park KC, Hetling JR. Bioelectronic Artificial Nose Using Four- Channel Moth Antenna Biopotential Recordings. Proceedings of the 2 International IEEE EMBS Conference on Neural Engineering, Arlington, Virginia; March 16 - 19, 2005;1-4.

[71] Zenina MA, Titov AV, Sarach OB, Guljaev AM, Muchina OB, Varlashov IB. On the Road to the Artificial Nose. 2nd SIBERIAN RUSSIAN STUDENT WORKSHOP ON EDM , SECTION I, ERLAGOL. 3-7 July,2001; 33-35.

[72] Hauptmann P, Borngraeber R, Schroeder J, Auge J. Artificial electronic tongue in comparison to the electronic nose - state of the art and trends. IEEUEIA International Frequency Control Symposium and Exhibition. 2000.

[73] Harrison LM.The Pocket Medical Dictionary. ISBN: 81-239-0926-8: 3: 419.

[74] *Hauptmann P, Borngraeber R, Schroeder J, Ifak JA*. Artificial electronic tongue in comparison to the electronic nose - state of the art and trends. *IEEE/EIA International Frequency Control Symposium and Exhibition. 2000;* 22-29.

[75] Hauptmann P . Sensors-principles and applications. Prentice Hall, Hertfordshire (UK).1993.

[76] DiNatale C, Davide F, D"Amico A, Legin A, Rudinitskaya A, Selezenev BI, Vlasov Y. Applications , of an electronic tongue to

the environmental control. Proc. EUROSENSORS X**,** Leuven (Belgium). 1996; 1345-48.

[77] Legin A, Rudinitskaya A, Vlasov Y, DiNatale C, Davide F, D"Amico A. Tasting of beverages using an electronic tongue", Sensors and Actuators. 1997; 291-96.

[78] Toko K .Taste sensor with global selectivity . Materials Science and Engineering. 1996; 69-82.

[79] Toko K. Taste sensor. Proc. Transducers. Sendei (Japan).1999; 58-61.

[80] Hauptmann P, Auge J, Borngraber R, Schroder J. Application of novel sensor electronics for quartz resonators in artificial tongue. Proc. IEEE/EIA Int. Frequency Control Symposium, Kansas City(USA). 2000, in press.

[81] Rabe, Biittgenbach S, Zimmermann B, Hauptmann P .Design, manufacturing, and characterization of high-frequency, thickness hear mode resonators. Proc. IEEE/EIA Int. Frequency Control Symposium, Kansas City (USA). 2000, in press.

[82] Gibbs WW. A tongue for love. Scientific American: Technology and Business. May 98.

[83] Yoo SJ**,** Lavigne J, Savoy S, McDoniel JB, Anslyn EV, McDevitt JT, Neikirk DP, Shear JB . Mikroma chined storage wells for chemical sensing beads in an ,artificial tongue. Proc. SPIE, Austin(USA). 1997;3224.

[84] Winquist F, Wide P, Lundstrom I . An electronic tongue based on voltammetry. Analytical Chimica Acta. 1997; 357: 21-31.

[85] Krantz-Riilcker C, Winquist F, Ekedahl LG . A new project area wth in the activities of **S**-Sense- Optimisation of processes by multi sensing technology. Proc. ISOEN **,** Tubingen (Germany). 1999; 61-64.

[86] Wide P, Winquist F, Bergsten P, Petriu EM. The Human-Based Multisensor Fusion Method for Artificial Nose and Tongue Sensor Data. IEEE Transactions on instrumentation and measurement. 1998; 47: 1072-77.

[87] Suganuma A, Kataoka M, Araki K. Application of a Neural Network to Human Tasting. IEEE 0730-3157192.1992; 277-282.

[88] Iiyama S, Toko K , Yamafuji K. Taste Reception in a Synthetic Lipid Membrane. *Maku (Membrane),1987; 12: 231-37.*

[89] Hayashi K, Yamafuji K, Toko K, N. Ozaki, T. Yoshida, S. Iiyama and N. Nakashima.Effect of Taste Substances on Electric Characteristics of a Lipid Cast Membrane with a Single Pore. Sensor and Actuators,,1989; 16**:** 25-42.

[90] Hayashi K, Yamanaka M, Toko K , Yamafuji K.Multichannel Taste Sensor Using Lipid Membranes.*Sensor and Actuators B.1990;*2: 205-13.

[91] Toko K, Hayashi K, Yamanaka M , Yamafuji K. Multi-Channel Taste Sensor with Lipid Membranes. *Technical Digest of the 9-th Sensor Symposium,1990;* 193-96, 1990.

[92] DeSiyne JA, Heck L, Mierson S, DeSimone SK. Canine lingual epithelia in vitro: for gustatory transduction. J Gen Phvsiol, The active ion transport properties of implications,1984; 83:633-56.

[93] Simon SA, Garvin JL. Salt and acid studies on canine lingual epithelium. Am J Phvsiol.1985; 249:C398-C408.

[94] Simon SA, Robb R,. Schiffmy SS. Transport pathways in rat lingual epithelium. Pharmacol Biochem Behav .1988;29:257-67.

[95] Mierson S, Heck L, DeSimone SK, Biber TUL. A current carriers in canine lingual epithelium in vitro. Biochim Biophys Acta 1985;816:283-93.

[96] Roper SD .The cell biology of the taste receptor. Ann. Rev. Neurosci.1989 ;12:329-53.

[97] DeSimone JA, Heck GL, Ye Q. Electrophysiological Stuoies Of Salt-Sensitive Taste Receptors. IEEE Ch2998-3/!9 1/0000-0286; 286-287.

[98] Toko K. Electronic Sensing of the Taste of Beer and Other Foodstuffs. IEEE 0-7803-2700-4; 6.3.1-6.3.6.

[99] Costa de Sousa H, Carvalho ACPLF, Riul A Jr, Mattoso LHC. Using MLP Networks to Classify Red Wines andWater Readings of an Electronic Tongue. Proceedings of the VII Brazilian Symposium on Neural Networks (SBRN"02) IEEE,2002.

[100] Freeman C, Dony RD , Areibi SM. Audio Environment Classification for Hearing aids using Artificial Neural networks with Windowed Input. Proceedings of the 2007 *IEEE Symposium on Computational Intelligence in Image and Signal Processing* (CIISP 2007). 183-86.

[101] Plomp R. Auditory handicap of hearing impairment and the limited benefit of hearing aids. J. Acoust. Soc. Am.1978; 63:533 –49.

[102] Nordqvist P , Leijon A. An efficient robust sound classification algorithm for hearing aids. J. Acoust Soc Am.2004; 115: 3033– 41.

[103] Buchler M, Allegro S, Launer S, Dillier N. Sound classification in hearing aids inspired by auditory scene analysis. EURASIP J. App. Sig. Pro.2005; 2005: 2991 – 3002.

[104] Haykin SS. Neural networks: a comprehensive foundation. New York: Macmillan, 1999.

[105] Rabiner L. A tutorial on hidden markov models and selected applications in speech recognition. *Proceedings of the IEEE*. 1989; 77, 2: 257 – 86.

[106] Khan MKS, Al-Khatib WG, Moinuddin M. Automatic classification of speech and music using neural networks. MMDB 2004: Proceedings of the Second ACM International Workshop on Multimedia Databases.2004; 94 – 9.

[107] Zongker D, Jain A. Algorithms for feature selection: An evaluation. Proceedings of the 13th International Conference on Pattern Recognition.1996; 2: 18– 22, 1996.

[108] Lee JM, Choi JS, Lim YS, Kim HS, Park M. Intelligent and Active System for Human-Robot Interaction based on Sound Source Localization. *International Conference on Control, Automation and Systems* in COEX, Seoul, Korea .Oct. 14-17, 2008. 2738-41.

[109] Vermaak J, Gangnet M, Blake A, Perez P. Sequential Monte Carlo fusion of sound and vision for speaker tracking. Proceedings of *IEEE* ICCVM Vancouver, July 2001.

[110] Pavlovic V, Garg A, Rehg J. Multimodal speaker detection using error feedback dynamic Bayesian networks. Proceedings of the IEEE CVPR, Hilton Head Island, SC, 2000.

[111] Ozdamar O, Lopez CN, Delgado RE. A Digital Speech Processor with Lateral Inhibition For Artificial Hearing. *IEEE* Engineering In Medicine & Biology Society 10th Annual International Conference. 1988; 1545-46.

[112] Yamada Y et all.. Slip phase isolating : impulsive signal generating vibrotactile sensor and its application to real-time object regrip control, Robotica.2000 ;18:.43-49.

[113] Johansson RS et.al.. Tactile sensory cording in the glabrous skin of the human hand. Trends *in* Neuro Sciences ,1983;6, 1: 27-32.

[114] Canepa G et.al. Detection of Incipient Object Slippage by Skin-Like Sensing and Neural Network Processing. *IEEE* Transactions *on* Systems, Man and Cybernetics-Part *B:* Cybernetics, 1998; 28, 3: 348-56.

[115] Trembly MR et.al.. Estimating friction using incipient slip sensing during a manipulation task. *IEEE* International conference *on* robotics and automation. 1993;429-34.

[116] Maeno T et al.. FE Analysis of the Dynamic Characteristics of the Human Finger Pad in Contact with Objects with/without Surface Roughness, Proc. 1998 *ASME* International Mechanical Engineering *Congress and* Exposition, DSC,1998; 64:279-86.